\documentclass{article}

\usepackage{arxiv}

\usepackage[utf8]{inputenc} 
\usepackage[T1]{fontenc}    
\usepackage{hyperref}       
\usepackage{url}            
\usepackage{booktabs}       
\usepackage{amsfonts}       
\usepackage{nicefrac}       
\usepackage{microtype}      
\usepackage{lipsum}
\usepackage{graphicx}
\usepackage[section]{placeins}
\usepackage{cite}
\usepackage{float}

\graphicspath{ {./images/} }

\title{Impact of Interventional Policies Including Vaccine on COVID-19 Propagation and Socio-Economic Factors}

\author{
  Haonan Wu \\
  Financial Innovation Labs (FinLabs)\\
  Synechron Inc. \\
  London EC2V 7NA, UK \\
  \texttt{haonan.wu@synechron.com} \\
   \And
  Rajarshi Banerjee \\
  Financial Innovation Labs (FinLabs)\\
  Synechron Inc. \\
  Bengaluru, Karnataka 560103, India \\
  \texttt{rajarshi.banerjee1@synechron.com} \\
   \And
  Indhumathi Venkatachalam \\
  Financial Innovation Labs (FinLabs)\\
  Synechron Inc. \\
  Bengaluru, Karnataka 560103, India \\
  \texttt{indhumathi.v@synechron.com} \\
   \And
  Daniel Percy-Hughes \\
  Financial Innovation Labs (FinLabs)\\
  Synechron Inc. \\
  London EC2V 7NA, UK \\
  \texttt{daniel.percy-hughes@synechron.com} \\
   \And
  Praveen Chougale \\
  Financial Innovation Labs (FinLabs)\\
  Synechron Inc. \\
  Bengaluru, Karnataka 560103, India \\
  \texttt{praveen.chougale@synechron.com} \\  
}

\begin{document}
\maketitle

\begin{abstract}
A novel coronavirus disease has emerged (later named COVID-19) and caused the world to enter a new reality, with many direct and indirect factors influencing it. Some are human-controllable (e.g. interventional policies, mobility and the vaccine); some are not (e.g. the weather). We have sought to test how a change in these human-controllable factors might influence two measures: the number of daily cases against economic impact. If applied at the right level and with up-to-date data to measure, policymakers would be able to make targeted interventions and measure their cost. This study aims to provide a predictive analytics framework to model, predict and simulate COVID-19 propagation and the socio-economic impact of interventions intended to reduce the spread of the disease such as policy and/or vaccine. It allows policymakers, government representatives and business leaders to make better-informed decisions about the potential effect of various interventions with forward-looking views via scenario planning. We have leveraged a recently launched open-source COVID-19 big data platform and used published research to find potentially relevant variables (features) and leveraged in-depth data quality checks and analytics for feature selection and predictions. An advanced machine learning pipeline has been developed armed with a self-evolving model, deployed on a modern machine learning architecture. It has high accuracy for trend prediction (back-tested with r-squared) and is augmented with interpretability for deeper insights.
\end{abstract}


\section{Introduction}
Many factors can influence COVID-19 \cite{arc1} pandemic growth, directly or indirectly. Some are human-controllable (e.g. testing coverage, interventional policy and people mobility \cite{arc2,arc3,arc4,arc5}); some are not (e.g. environmental factors: temperature, wind speed and humidity \cite{arc6,arc7}).

We have sought to test how a change in these human-controllable factors might influence two measures: the number of daily cases against economic impact. If applied at the right level and with up-to-date data to measure, policymakers would be able to make targeted interventions and measure their cost. A visual description of our model's front-end and its usability is provided in [Multimedia Appendix 1].

Big data techniques always play a vital role in large-scale predictive analytics studies, especially considering the current COVID-19 pandemic situation, the quality and granularity for both historical and real-time data are critical. In this study, a high quality and open-source Big data platform is employed -the C3 AI COVID-19 Data Lake, which uniquely integrates multiple data sources from different geographies in a unified data model, ready for analysis \cite{arc8,arc9}.

\section{Methods}
\label{sec:headings}

\subsection{Data sources}
We started by using published research to understand relevant variables and leveraging the C3 AI COVID-19 Data Lake for the data collection and data access steps with quality checks to determine usability. We opted to use US State-level data as this had the broadest coverage for testing and would enable scaling of the approach to other jurisdictions.

In our analytics framework, we use environmental factors (including temperature, wind speed and humidity \cite{arc10}, testing count \cite{arc11}, interventional policies \cite{arc12}, mobility (location exposure) \cite{arc13}, and "day-of-week" as inputs, augmented with smart imputations from alternate sources and neighbouring data points for missing values and outliers. For example, for the confirmed daily cases in the U.S., we mainly use the data from Johns Hopkins University \cite{arc14} for its overall sound quality, but also impute negative and missing values by consolidating with other sources from C3 AI COVID-19 Data Lake. The high-frequency economic data for United States locations \cite{arc15} is used for the study of the socio-economic impact.
\subsection{Machine Learning Pipeline and Architecture}
We developed an ensemble machine learning model with auto/semi-auto hyperparameter tuning. The output predicts daily new cases and economic factors and allows for the simulation of interventions, including a vaccine. It always learns from the most recent data, so the model remains relevant as datasets evolve to show the change in people's behaviour.  For example, the current model in production was trained on the dataset until November 12, 2020. This framework is automatically scheduled to retrain the model monthly to catch up with the new data, always to provide up-to-date prediction and simulation. 

The user interface allows simulation by changing human-controllable input factors and recalculating predictions for a 5-week horizon.

The ensembled model pipeline is built in Scikit-learn, including advanced models, such as Linear Regression, Random Forest and Gradient Boosting \cite{arc16}. The hyperparameters are also fine-tuned. Lagging mechanics were created on time-aware metrics to enable our model to provide a medium-term trend prediction (Figure 1).

\begin{figure}[H]
\caption{Model pipeline and architecture diagram}
\centering
\includegraphics[scale=0.6]{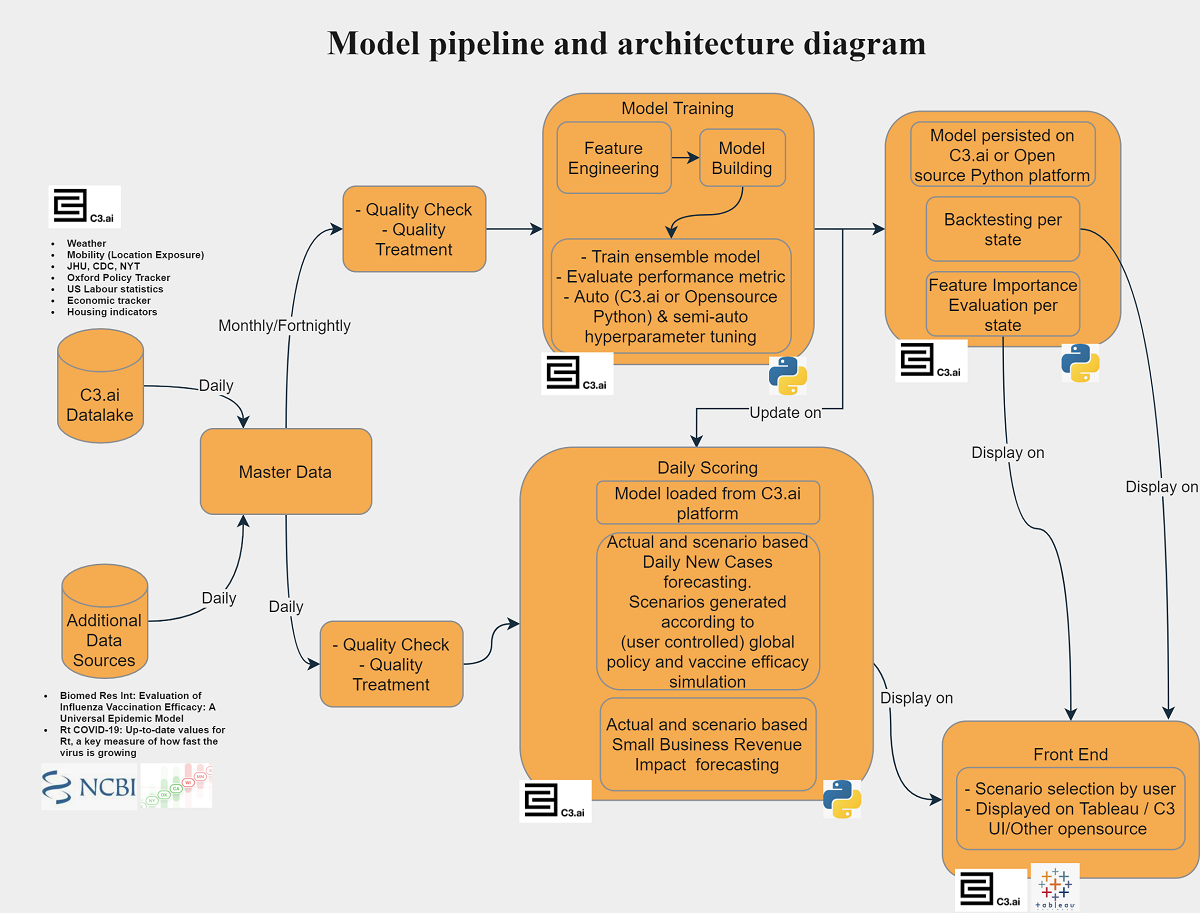}
\end{figure}

Good accuracy was obtained on many large U.S. states for medium-term trend predictions of daily cases based on back-testing (with the recent 14 days used as Out-of-Time testing set), we selected r-squared as the accuracy metric, since it measures the trend fitting (Figure 2).

\begin{figure}[h!]
\caption{Back-testing with r-squared as the metric for monitoring the trend prediction accuracy}
\centering
\includegraphics[scale=0.6]{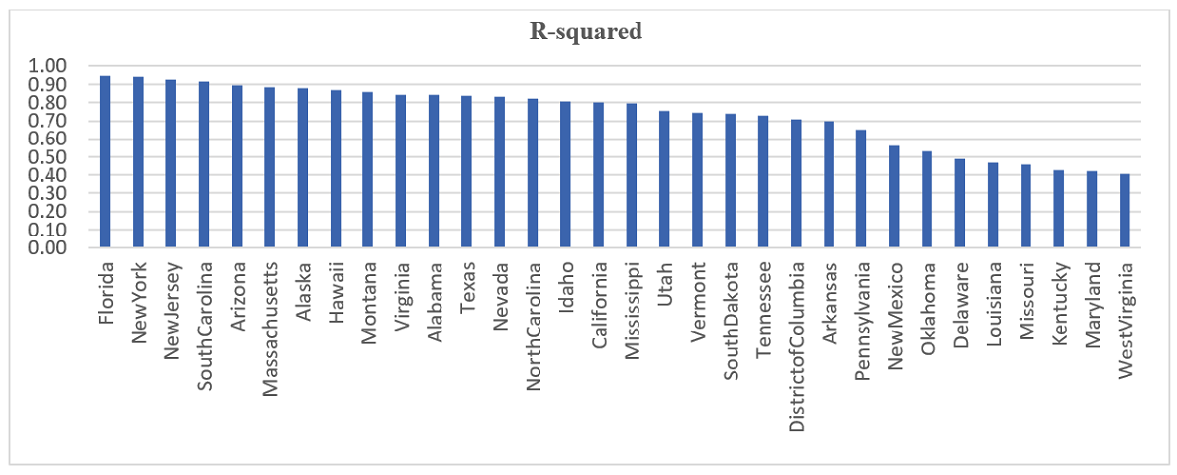}
\end{figure}

Several interpretability methods were experimented to explain the model outcome at both the global and local levels (e.g. Permute ranking \cite{arc17}, Partial dependence \cite{arc18}, LIME \cite{arc19} and SHAP \cite{arc20}).  We deployed a customized time-sensitive and cases weighted LIME method in our current graphical user interface.

The data and model from the influenza pandemic vaccination efficacy study \cite{arc21}, adopted with COVID-19's real-time reproduction rates \cite{arc22}, is used as the proxy for the future COVID-19 vaccine. This model is applied to mimic the situation when COVID-19 vaccines will be available. 

Our source code is accessible at [Multimedia Appendix 2].

\section{Results}
Taking the California model as an example, based on the recent model trained on the date of November 12, 2020, via prediction \& simulations, we can provide many valuable data-driven insights: 
\begin{itemize}
\item Based on the current policies, in a 5-week horizon, the virus spread shows an upward trend whilst small business revenue is downward. The 'Stay-at-home' restriction policy is the most impactful factor suggested by the model (Figure 3). A more detailed description of policy types and indices can be found from University of Oxford-Coronavirus Government Response Tracker \cite{arc23}.

\begin{figure}[h!]
\caption{The prediction of virus spread and small business revenue for California based on the current policy setting}
\centering
\includegraphics[scale=0.6]{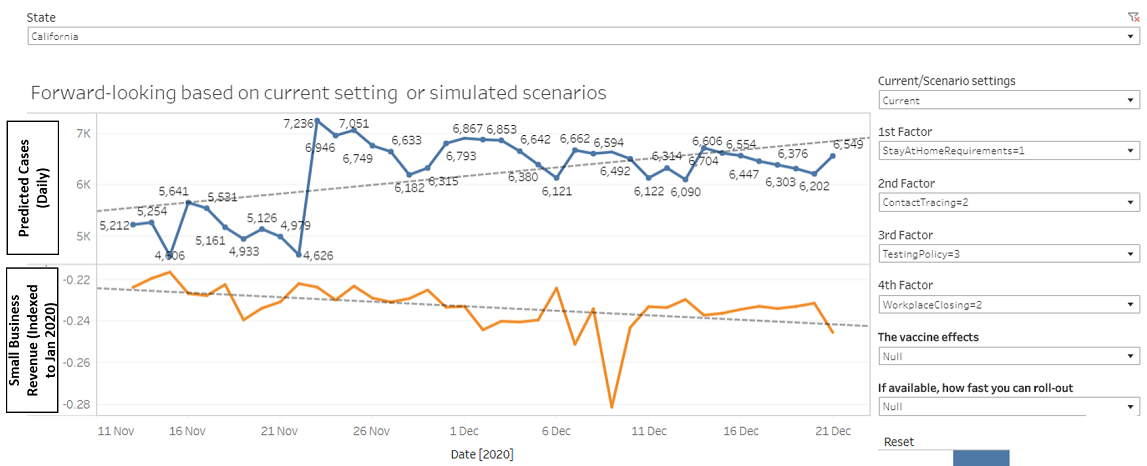}
\end{figure}

\item By moving the 'Stay-at-home' variable to a higher level (3), we can see the virus spread turning downward, but the small business revenue could also be negatively impacted (Figure 4). By moving Stay-at-home requirement to the lowest level (0), the small business revenue shows recovery; however, the virus spread is trending up (Figure 5). Thus, it is a dilemma that forces people to make compromises.

\begin{figure}[h!]
\caption{Simulate the virus spread and small business revenue for California with a higher level of 'Stay-at-home' restriction}
\centering
\includegraphics[scale=0.6]{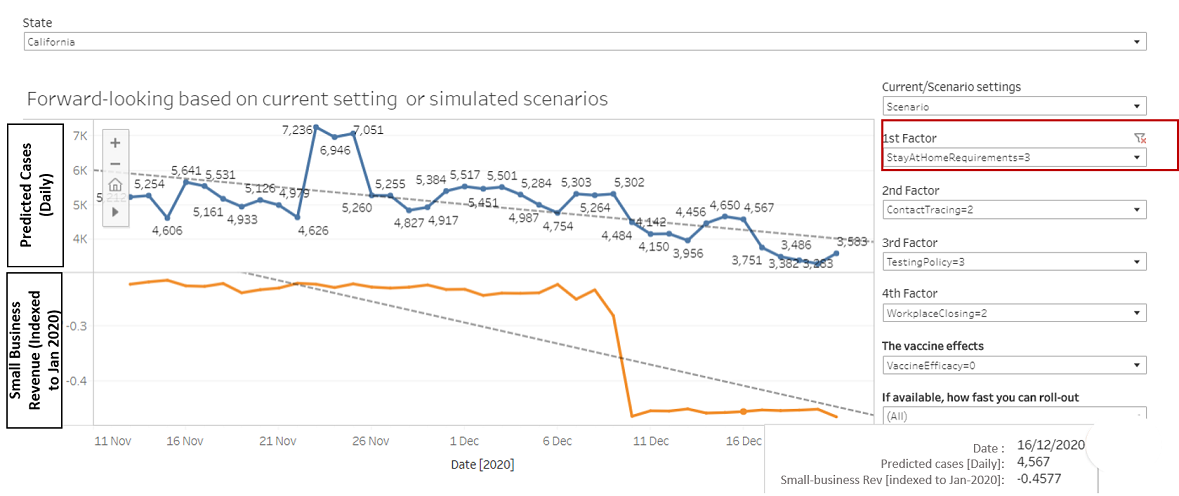}
\end{figure}

\begin{figure}[h!]
\caption{Simulate the virus spread and small business revenue for California with a lower level of 'Stay-at-home' restriction}
\centering
\includegraphics[scale=0.6]{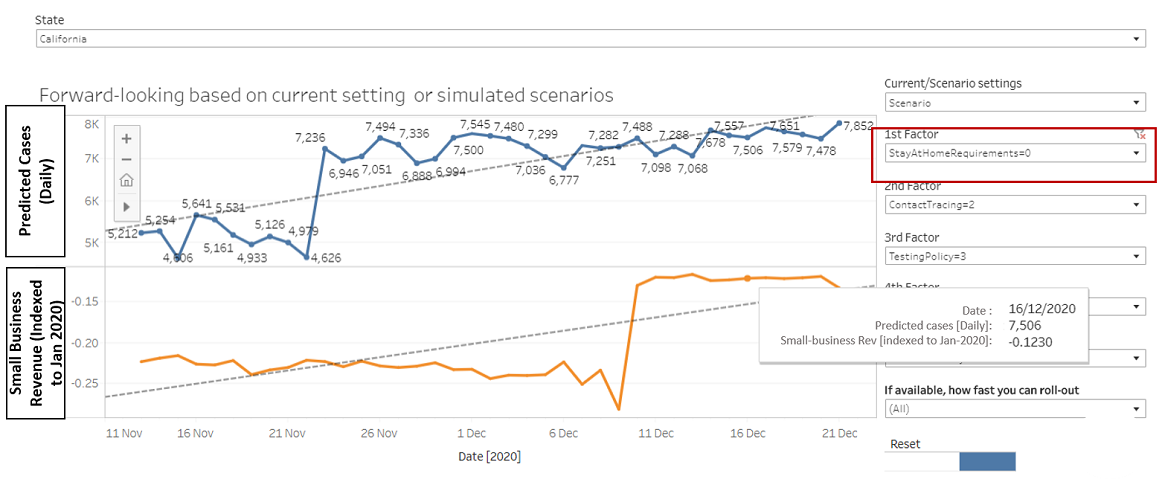}
\end{figure}

\item With a vaccine, we can see the possibility to both control the virus and recover the economy. Moreover, the user can simulate to find best-case scenarios, such as maximizing both the protect rate and small business revenue recovery level (Figure 6). When the COVID-19 vaccine will become available, together with more parameters about the different types of vaccines (e.g. Efficacy level, number of doses required, Price, Cost of delivery, roll-out speed, etc.) , the policymaker can use our model and its outputs for the vaccine selection and roll-out optimization.

\begin{figure}[h!]
\caption{Best-case scenarios simulation if the COVID-19 vaccines would become available}
\centering
\includegraphics[scale=0.6]{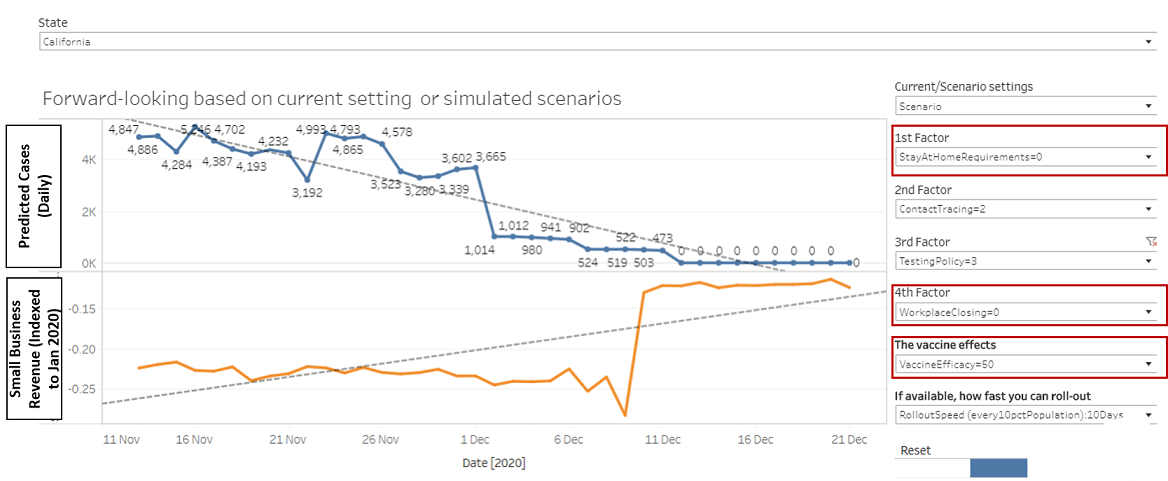}
\end{figure}
\end{itemize}

\section{Discussion}
Human behaviour and extreme natural disasters (like the COVID-19 pandemic) are hard to measure with data points. No model has the ability to provide an answer that is correct 100\% of the time, counter-intuitive results can always be found; however, with high-quality analytics and models with feedback mechanics, a forward-looking view can be inferred or at least noted.

In this paper, we applied the Machine Learning and Big Data techniques to model, predict and simulate COVID-19 daily-cases and the socio-economic impact of interventions intended to reduce the spread of the disease such as policy and/or vaccine. This is intended to help the government to test scenarios, plan proactive actions, optimize logistics and create a more open dialogue with the general public.  The predictive analytics model is deployed on a modern architecture that is scalable, self-evolving and explainable. Current models are at U.S. state-level and can be quickly rolled out to other countries/regions with datasets at a similar scale.

\section{Multimedia Appendix 1}
A visual description of our model's front-end and its usability \cite{arc24}.

\section{Multimedia Appendix 2}
The source code of our model \cite{arc25}.

\bibliographystyle{unsrt}  


\end{document}